\documentclass[preprint,12pt]{elsarticle}
\usepackage[tableposition=top]{caption} 
\usepackage[all]{nowidow}
\usepackage{xcolor}

\usepackage{amssymb}

\journal{Applied Soft Computing}

\begin{document}

\begin{frontmatter}



\title{Evaluating Named Entity Recognition: A Comparative Analysis of Mono- and Multilingual Transformer Models on a Novel Brazilian Corporate Earnings Call Transcripts Dataset}


\author[inst1,inst2]{Ramon Abilio\corref{cor1}}
\ead{ramon.abilio@ifsp.edu.br}
\cortext[cor1]{Corresponding author}

\author[inst2]{Guilherme Palermo Coelho}
\ead{gpcoelho@unicamp.br}

\author[inst2]{Ana Estela Antunes da Silva}
\ead{aeasilva@unicamp.br}

\affiliation[inst1]{organization={Instituto Federal de São Paulo - IFSP},
            addressline={Av. Zélia de Lima Rosa, 100 - Recanto das Primaveras I}, 
            city={Boituva},
            postcode={18552-252}, 
            state={São Paulo},
            country={Brazil}}

\affiliation[inst2]{organization={Universidade Estadual de Campinas - Unicamp},
            addressline={R. Paschoal Marmo, 1888, Jardim Nova Itália}, 
            city={Limeira},
            postcode={13484-332}, 
            state={São Paulo},
            country={Brazil}}

\begin{abstract}

Since 2018, when the Transformer architecture was introduced, Natural Language Processing has gained significant momentum with pre-trained Transformer-based models that can be fine-tuned for various tasks. Most models are pre-trained on large English corpora, making them less applicable to other languages, such as Brazilian Portuguese. In our research, we identified two models pre-trained in Brazilian Portuguese (BERTimbau and PTT5) and two multilingual models (mBERT and mT5). BERTimbau and mBERT use only the Encoder module, while PTT5 and mT5 use both the Encoder and Decoder. Our study aimed to evaluate their performance on a financial Named Entity Recognition (NER) task and determine the computational requirements for fine-tuning and inference. To this end, we developed the Brazilian Financial NER (BraFiNER) dataset, comprising sentences from Brazilian banks' earnings calls transcripts annotated using a weakly supervised approach. Additionally, we introduced a novel approach that reframes the token classification task as a text generation problem. After fine-tuning the models, we evaluated them using performance and error metrics. Our findings reveal that BERT-based models consistently outperform T5-based models. While the multilingual models exhibit comparable macro F1-scores, BERTimbau demonstrates superior performance over PTT5. In terms of error metrics, BERTimbau outperforms the other models. We also observed that PTT5 and mT5 generated sentences with changes in monetary and percentage values, highlighting the importance of accuracy and consistency in the financial domain. Our findings provide insights into the differing performance of BERT- and T5-based models for the NER task.

\end{abstract}

\begin{graphicalabstract}
    \includegraphics[width=\linewidth]{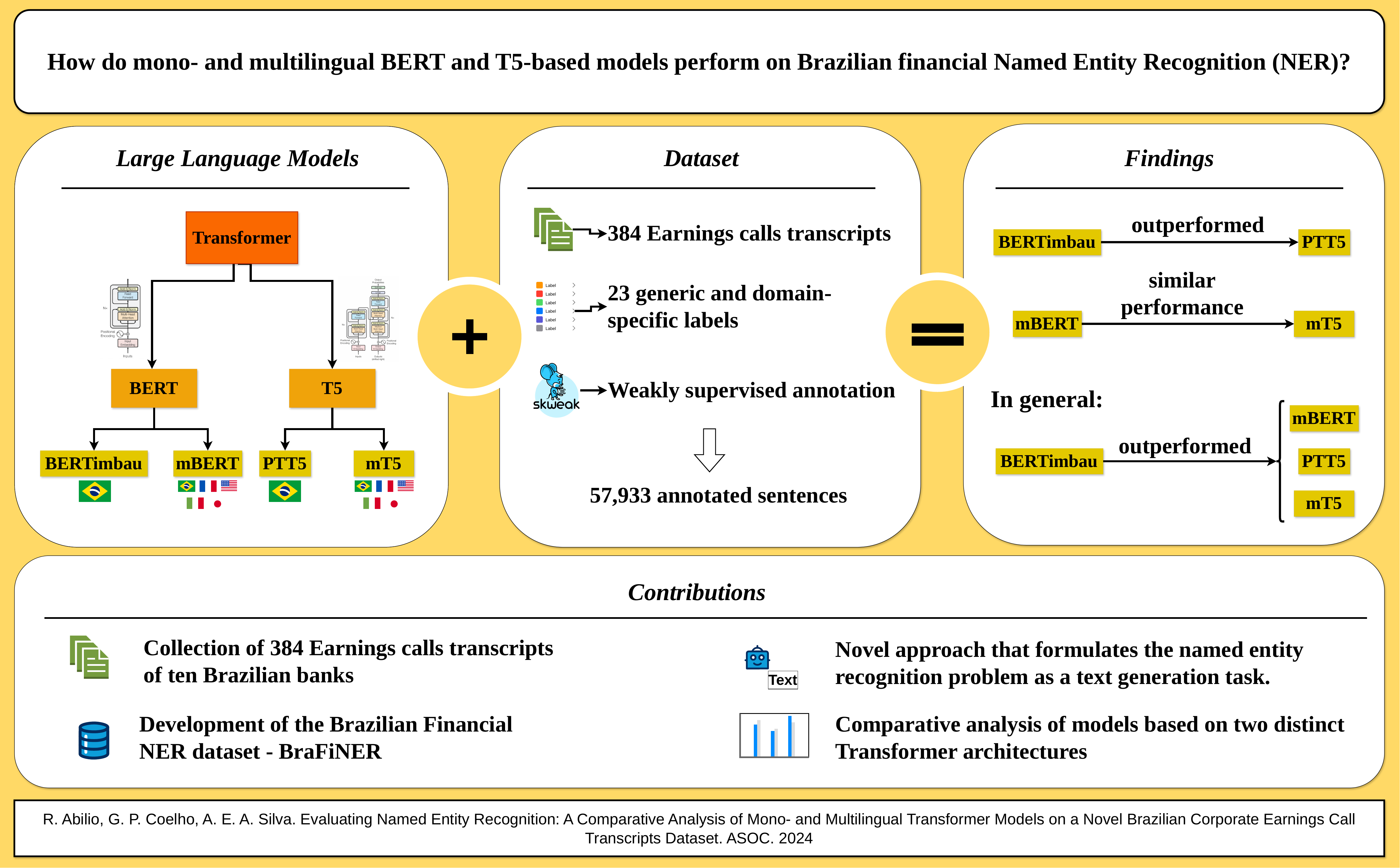}
\end{graphicalabstract}

\begin{highlights}
    \item Compilation of 384 conference call transcripts from Brazilian banks.
    \item Development of the BraFiNER, a Portuguese dataset for Named Entity Recognition in finance.
    \item Framing NER as text generation with T5, surpassing prior methods.
    \item Comparison of BERT and T5 models shows insights into their NER efficacy.
    \item The macro F1-score achieved by the models ranged from 98.33\% to 98.99\%.
\end{highlights}


\end{frontmatter}


\section{Introduction}
\label{sec:introduction}

In the investment field, various techniques are employed for stock selection in portfolios. These techniques encompass market trend analysis with Technical Analysis \cite{Siegel2014}, sentiment analysis of textual data from sources such as social media and news \cite{Carosia2021}, as well as fundamental analysis. Fundamental analysis involves scrutinizing historical financial metrics such as profit, revenue, liabilities, net earnings, cash flow, and more, alongside qualitative data found in press releases and quarterly or annual reports \cite{Prado2018}.

Automating the extraction of information from these reports could significantly aid in their analysis. Named Entity Recognition (NER), a Natural Language Processing (NLP) technique, plays a crucial role in this context by identifying entities such as people, places, companies, dates, and currencies within text \cite{Li2022}. These entities are then labeled accordingly; for example, the word ``London'' may be identified and labeled as a location.

NER serves as a vital information extraction technique and is often integrated with other NLP tasks such as summarization and question-answering \cite{Li2022}. Its versatility extends across diverse domains, including biomedical and financial sectors \cite{Alvarado2015, Francis2019, Song2021, Loukas2022, Zhang2023, Wang2023-FinGPT, Ogrinc2024, Krstev2024}, automatic speech recognition \cite{DelRio2021}, and legal contexts \cite{Araujo2018}. Regarding the application of NER in the financial domain, existing studies have primarily focused on generic entities such as Person, Location, Percentage, and Money \cite{Alvarado2015, Francis2019, DelRio2021, Loukas2022, Shah2023, Wang2023-FinGPT}.

Recognizing entities in texts can be achieved through various methods~\cite{Li2022, Song2021}. Traditional techniques rely on manually defined dictionaries and rules, which are contingent upon the quantity and quality of the defined words and rules. In contrast, machine learning techniques leverage pattern learning to identify entities in unseen texts, although they require extensive feature engineering and domain knowledge \cite{Song2021}. Alternatively, deep learning approaches have gained traction for entity recognition due to their ability to automatically uncover hidden features \cite{Li2022, Song2021}.

As a deep learning-based approach, the Transformer model architecture includes both Encoder and Decoder modules \cite{Vaswani2017}. Since its publication in 2018, various Transformer-based models have been pre-trained or fine-tuned for downstream tasks. For example, Bidirectional Encoder Representations from Transformers (BERT) \cite{Devlin2019}, Text-to-Text Transformer (T5) \cite{Raffel2020}, and Generative Pre-Training (GPT) \cite{Radford2018-gpt} are models based on the Transformer architecture, but they differ in their structure. While BERT is an Encoder-based model, T5 employs an Encoder-Decoder architecture, and GPT is a Decoder-based model. These models have demonstrated efficacy in NER tasks \cite{Li2022, Francis2019, Song2021, Zhang2023, Wang2022instructionner, Wang2023gptner, Wang2023-FinGPT, Ogrinc2024, Krstev2024}. However, their application has been predominantly centered on English texts, overlooking the specific challenges and nuances present in other languages.

Given the importance of NER in the financial domain, it is essential to explore approaches to entity recognition in non-English texts, such as Brazilian Portuguese, using different model architectures. In this study, we aim to perform a comparative analysis of models with different Transformer-based architectures pre-trained in Portuguese. We compared the performance of BERT and T5-based models in the financial NER task, assessed the computational resources required for fine-tuning, and analyzed their errors in recognizing entities in financial texts written in Portuguese.

We encountered several challenges during this process. The first challenge was finding models with different architectures pre-trained in Portuguese. We selected two monolingual models pre-trained in Portuguese (BERTimbau \cite{Souza2020} and PTT5 \cite{Carmo2020}) along with two multilingual models (mBERT \cite{Devlin2019} and mT5 \cite{Xue2021}). The second challenge was related to the T5 architecture. Since T5 is a generative model, we had to adapt the token classification task as used in BERT to a sequence-to-sequence approach. The third challenge was the lack of a suitable dataset. We did not find a dataset based on financial texts from companies listed on the Brazilian stock exchange annotated for named entity recognition. Therefore, we prepared the Brazilian Financial NER Dataset (BraFiNER) using 384 earnings call transcripts from ten Brazilian banks and used it for model fine-tuning and evaluation.

The main contributions of this work are outlined as follows:

\begin{itemize}
    \item Compilation of a comprehensive collection comprising 384 conference call transcripts from ten prominent Brazilian banks. This dataset encompasses approximately 118,411 sentences and 3,082,526 tokens, facilitating robust analysis and experimentation.

    \item Development of the Brazilian Financial NER Dataset (BraFiNER) containing sentences annotated using weakly supervised techniques. This dataset serves as a valuable resource for training and evaluating named entity recognition (NER) systems in the financial domain.

    \item Proposal of a novel approach that formulates the named entity recognition problem as a text generation task. This innovative perspective offers insights into alternative methodologies for addressing NER challenges and has demonstrated superior performance using a T5 model with fewer parameters compared to existing approaches.

    \item Comparative analysis of results obtained from models based on two distinct Transformer architectures: Encoder-only (BERT) and Encoder-Decoder (T5). This analysis provides valuable insights into the relative efficacy and performance of these architectures in the context of NER tasks.
\end{itemize}

\section{Related Work}
\label{sec:related_work}

Named Entity Recognition (NER) has been extensively studied over the years using various datasets, label sets, and strategies \cite{Li2022, Song2021}. The following subsections present related works on the pre-training of Transformer-based models for the financial domain, the fine-tuning of Transformer-based models, and the development of datasets for financial NER.

\subsection{Pre-training Transformer-based models for the Financial domain}

The application of pre-trained models in the financial domain has been extensively studied over the years \cite{Li2023-LLM-Finance-Survey, Lee2024-LLM-Finance-Survey}, and various Transformer-based models have been developed \cite{Lee2024-LLM-Finance-Survey}. Examples of these models include FinBERT \cite{Liu2021-FinBERT}, FinBERT PT-BR \cite{Santos2023-FinBERT-PTBR}, and FLANG-BERT and FLANG-ELECTRA~\cite{Shah2022-Flue}.

FinBERT \cite{Liu2021-FinBERT} is a BERT-based model pre-trained from scratch using both general and financial English corpora. The FinBERT-PT-BR \cite{Santos2023-FinBERT-PTBR} model is based on BERTimbau \cite{Souza2020}, another BERT-based model, but pre-trained on Brazilian Portuguese corpora. In FinBERT-PT-BR, the authors continued the pre-training of BERTimbau by adding news from the Brazilian financial market. Shah et al. \cite{Shah2022-Flue} released two models: FLANG-BERT, based on the BERT architecture, and FLANG-ELECTRA, based on the ELECTRA architecture. These models were pre-trained on general and financial English datasets.

In these examples, the authors pre-trained their models and fine-tuned them for downstream tasks such as Sentiment Analysis and NER. Unlike these works, we fine-tuned BERT- and T5-based models that were either pre-trained exclusively in Brazilian Portuguese (monolingual) or included Brazilian Portuguese in their corpora (multilingual) for the NER task. Besides, unlike Santos et al. \cite{Santos2023-FinBERT-PTBR}, our dataset comprises text from earnings call transcripts for NER, while they used financial news for Sentiment Analysis.

\subsection{Fine-tuning of Transformer-based models for Financial NER}

Regarding the fine-tuning of Transformer-based models for the NER task, BERTimbau \cite{Souza2020}, a Portuguese pre-trained BERT-based model, achieved state-of-the-art performance on the Portuguese MiniHAREM dataset. Similarly, PTT5 \cite{Carmo2020}, introduced in 2020, is a Portuguese pre-trained T5 model. After pre-training, it was fine-tuned for various tasks, including NER. In their NER experiments, the authors prefixed the instruction ``Recognize Entities'' to sentences, as in ``Recognize Entities: John lives in New York''. They expected an output following the pattern: ``John [Person] lives in [Other] New York [Location]'', where the words in brackets represent the entity types. Experiments were conducted using the Portuguese datasets HAREM and MiniHAREM, widely utilized for NER tasks, resulting in an F1-score of 82\%.

In 2021, Finardi et al. \cite{Finardi2021} introduced BERTaú, a BERT-based model pre-trained from scratch using data from the Brazilian bank Itaú virtual assistant. This model was fine-tuned for various tasks including Frequent Answer Question Retrieval, Sentiment Analysis, and NER. Their NER dataset, which comprises 18,370 manually annotated examples across 16 classes including specific banking entities, yielded an F1-score of 0.877 after 5 epochs of fine-tuning. Additionally, they reported F1-scores of 0.840 for mBERT uncased and 0.853 for BERTimbau base. During inference, a V100 GPU was employed, with an average processing time of 0.01944 seconds per sample.

In 2022, Wang et al. \cite{Wang2022instructionner} introduced the InstructionNER framework, which leverages the T5-large model to address the Named Entity Recognition (NER) task by converting it from token classification to a sequence-to-sequence (seq2seq) format. Fundamentally, the approach involves feeding the model with a sentence, an instruction, and options. The sentence represents the text in which named entities are to be identified, the instruction guides the model on what action to take, and the options denote the types of entities. The model's output adheres to the template ``entity is a/an type''. The authors employed few-shot NER for fine-tuning, utilizing five datasets distinct from the financial domain. Across their experiments, they achieved a maximum F1 value of 0.954.

In 2024, Krstev et al. \cite{Krstev2024} fine-tuned multilingual BERT-based models to perform NER and Relation Extraction on analysts stock rating news. They obtained 180,000 financial-related texts written in English and German, filtered those related to analysts ratings, and manually labeled about 1000 samples. For NER, they considered eight classes related, for example, to the analyst and company's name, the price target, and the position of the analyst.

Similar to the work of Carmo et al. \cite{Carmo2020}, Finardi et al. \cite{Finardi2021}, Wang et al. \cite{Wang2022instructionner}, and Krstev et al. \cite{Krstev2024}, our study involved fine-tuning BERT- and T5-based models for the Named Entity Recognition (NER) task. However, our approach diverged from that of Carmo et al. \cite{Carmo2020}, who introduced PTT5—a T5 model pre-trained from scratch and subsequently fine-tuned for multiple tasks. In contrast, we fine-tuned their pre-trained model and the mT5 model specifically for domain-specific NER. Additionally, while Finardi et al. \cite{Finardi2021} pre-trained BERT from scratch and fine-tuned it for a variety of tasks, our work was focused on fine-tuning and comparing two BERT-based models pre-trained on Portuguese (BERTimbau) and multilingual (mBERT) datasets, exclusively for NER. Furthermore, our study emphasized domain-specific entities such as profit, revenue, net income, and cash flow, whereas Finardi et al.'s \cite{Finardi2021} work was centered on entities related to customer services. Finally, while Krstev et al. \cite{Krstev2024} concentrated on multilingual BERT-based models applied to news regarding analysts' stock ratings written in English and German, with a focus on specific entities, we compared monolingual and multilingual models using different Transformer architectures for texts written in Brazilian Portuguese, targeting a broader range of entities suitable for various applications.

In contrast to the approach of Wang et al. \cite{Wang2022instructionner}, who employed the large version of T5, we fine-tuned two T5-based models with fewer parameters pre-trained on Portuguese (PTT5 base version) and multilingual (mT5 small version) datasets. Similarly, comparable to the methodologies of Carmo et al. \cite{Carmo2020} and Wang et al. \cite{Wang2022instructionner}, we reformulated the NER task from token classification to seq2seq. However, our approach diverges from theirs in that we fine-tuned the models using a sentence as input and generated the same sentence as output, with entities and their labels annotated within the sentence.

\subsection{Dataset development for Financial NER}

In 2015, Alvarado, Verspoor, and Baldwin \cite{Alvarado2015} constructed a NER dataset utilizing the CoNLL-2003 dataset available in the literature, in addition to a supplementary dataset created by them, which comprised domain-specific English texts. This supplementary dataset included texts extracted from financial agreements, sourced from 8 documents (totaling 54,256 words) randomly selected for manual annotation. They annotated four types of entities provided in the CoNLL-2003 dataset: LOCATION (LOC), ORGANIZATION (ORG), PERSON (PER), and MISCELLANEOUS (MISC). Furthermore, they automatically annotated all instances of ``lender'' and ``borrower'' as PERSON (PER). Experiments were conducted using various combinations of training and testing datasets, employing Conditional Random Fields (CRFs) for annotation. The highest F-score (82.7\%) was attained when the CRF was trained and tested with the domain-specific dataset.

In 2019, Francis, Van Landeghem, and Moens \cite{Francis2019} developed a dataset using 3000 invoices from service providers such as insurance, telecommunications, banking, and tax companies. They employed optical character recognition (OCR) to extract English texts related to International Bank Account Number (IBAN) of the beneficiary, invoice number, invoice and due dates, total inclusive amount, total exclusive amount, total Value Added Tax (VAT) amount, structured communication, company name, company address, and other related fields required for invoice processing. The NER task was performed using an architecture consisting of an input layer, two stacked Bidirectional Long Short-Term Memory (Bi-LSTM), and a CRF layer. Due to variances in entity structure between the gold dataset\footnote{A gold dataset is considered to be of high quality and accuracy due to the meticulous manual annotation process conducted by human annotators.} and the extracted entities, the authors evaluated their strategy manually.

Two years later, in 2021, Del Rio et al. \cite{DelRio2021} compiled a dataset using English transcripts from 44 earnings calls across nine sectors, recorded in 2020. They meticulously annotated 24 types of entities, including PERSON, ORG, LOC, PRODUCT, EVENT, WORK\_OF\_ART, LAW, LANGUAGE, DATE, TIME, PERCENT, MONEY, QUANTITY, ORDINAL, CARDINAL, ABBREVIATION, CONTRACTION, ALPHANUMERIC, YEAR, and WEBSITE. The annotation proceeded in three steps: 1) Their custom tool was used to annotate tokens requiring normalization, such as abbreviations, cardinals, ordinals, and contractions; 2) SpaCy 2.3.5\footnote{https://spacy.io/} was employed to annotate, for example, ORG and PERSON; and 3) Manual review of annotations and updates were conducted.

Subsequently, in 2022, Loukas et al. \cite{Loukas2022} introduced the FINER-139 data\-set. This extensive dataset contains 1.1 million sentences tagged with gold XBRL (e{X}tensible Business Reporting Language) labels. Extracted from reports of publicly traded companies filed with the US Securities and Exchange Commission (SEC) over five years (2016 to 2020), it encompasses the 139 most prevalent XBRL entity types among nearly 6,000 entity types. The authors conducted experiments employing Bi-LSTMs and BERT to evaluate the dataset's utility and applicability.

Shah et al. \cite{Shah2023}, in 2023, introduced a weakly supervised NER pipeline for financial texts, named FiNER, and created the FiNER-ORD dataset. This dataset comprises manually annotated financial news articles in English. From a pool of nearly 48,000 news articles, the authors selected 201 and annotated them using the Doccano\footnote{https://doccano.github.io/doccano/} software. The annotations focused on generic entities such as PERSON, LOCATION, and ORGANIZATION. Subsequently, the annotated dataset was utilized to validate the efficacy of the developed pipeline.

Similarly, in 2023, Zhang et al. \cite{Zhang2023} conducted experiments involving large language models and the recognition of named entities in the Chinese financial domain. They constructed a dataset primarily consisting of text sourced from financial websites, company earnings releases, financial short messages on social media, and financial article summaries, totaling 7,521 sentences. Manual annotation was performed on these sentences, covering six types of generic entities: PERSON\_NAME, PRODUCT\_NAME, COMPANY\_NAME, LOCATION, ORG\_NAME, and TIME.

In line with previous research efforts, we have curated a dataset tailored for the NER task. However, our dataset comprises Portuguese text sourced from 384 earnings call transcripts of 10 banks listed on the Brazilian stock exchange. Employing weak supervision, we annotated the sentences, incorporating both generic and domain-specific labels.

\section{Background}
\label{sec:background}

The Transformer is a model architecture composed by an encoder-decoder structure \cite{Vaswani2017}. The encoder receives a sequence of symbols as input and maps them to a sequence of continuous representations. Based on this representation, the decoder generates an output sequence, one element at a time in an auto-regressive fashion. The Transformer structure is flexible and has been adapted for different purposes and architectures. Two examples of these adaptations are: the Bidirectional Encoder Representations from Transformers (BERT) \cite{Devlin2019}, which uses only the encoder structure, and the Text-to-Text Transfer Transformer (T5) \cite{Raffel2020}, based on the encoder-decoder structure. BERT and T5 can be pre-trained and fine-tuned to different tasks.

\subsection{BERT-based models: mBERT and BERTimbau}

Devlin et al. \cite{Devlin2019} introduced BERT in 2019. This model utilizes only the Transformer's Encoder and was pre-trained using bidirectional (left and right context) representations from unlabeled text, with WordPiece as its tokenizer~\cite{Devlin2019}. BERT can be fine-tuned with an additional output layer for various tasks, including question answering, language inference, and token classification~\cite{Devlin2019}. BERT utilizes a fixed positional encoding scheme, which imposes a limitation on the input length. 

BERT was pre-trained only in English \cite{Devlin2019}. However, the authors provided multilingual models (mBERT) on Hugging Face\footnote{https://huggingface.co/}: BERT-base multilingual cased and BERT-base multilingual uncased. These models were pre-trained with Wikipedia texts from 104 different languages and the authors recommend the version with 12 layers and 110M parameters. 

Despite the introduction of mBERT, Souza, Nogueira and Lotufo presented BERTimbau in 2020 \cite{Souza2020}. This model, built on the BERT architecture, was pre-trained from scratch using BrWaC (Brazilian Web as Corpus) dataset. The authors released BERTimbau on Hugging Face in two versions: BERTimbau-base (12 layers and 110M parameters) and BERTimbau-large (24 layers and 335M parameters).

Typically, BERT models have a constraint of 512 tokens in their input, including the special tokens ``[CLS]'' and ``[SEP]''. Therefore, input sequences longer than 512 tokens must be truncated or split into smaller segments to comply with this limit. Notably, mBERT and BERTimbau models available on Hugging Face also adhere to this constraint.

\subsection{T5-based models: PTT5 and mT5}

The Text-to-Text Transfer Transformer (T5) model is based on the Transformer encoder-decoder architecture and uses SentencePiece as its tokenizer \cite{Raffel2020}. To use T5, we need to map our problem to a text-to-text format, i.e. taking text as input and generating text as output \cite{Raffel2020}. T5 was pre-trained using ``Colossal Clean Crawled Corpus'' (C4) and it was evaluated on a wide variety of NLP tasks such as question answering, document summarization, and sentiment classification \cite{Raffel2020}. Instead of using a fixed embedding for each position like BERT, T5 uses relative position embeddings. 

Similar to BERT, T5 has also served as a base for other models. Carmo et al. \cite{Carmo2020} introduced, in 2020, a model based on T5 architecture and pre-trained from scratch using BrWaC (Brazilian Web as Corpus) dataset. The authors named this model PTT5 and released three versions on Hugging Face: Small (60M parameters), Base (220M parameters), and Large (740M parameters). They also recommend the use of the Base version.

A multilingual version of T5 was introduced by Xue et al. \cite{Xue2021} in 2021. The model, named mT5, was pre-trained using a multilingual version of C4 (mC4 - 101 languages). mT5 was released in five versions: mT5-Small (300M parameters), mT5-Base (580M parameters), mT5-Large (1.2B parameters), mT5-XL (3.7B parameters), and mT5-XXL (13B parameters).

T5 models can process longer inputs than BERT models. However, longer inputs may still lead to increased computational cost and memory usage during training and inference. We confirmed that PTT5 and mT5 provided on Hugging Face have an input length of 512.

\section{Methodology}
\label{sec:methodology}

\begin{figure}[!ht]
    \centering
    \includegraphics[scale=0.33]{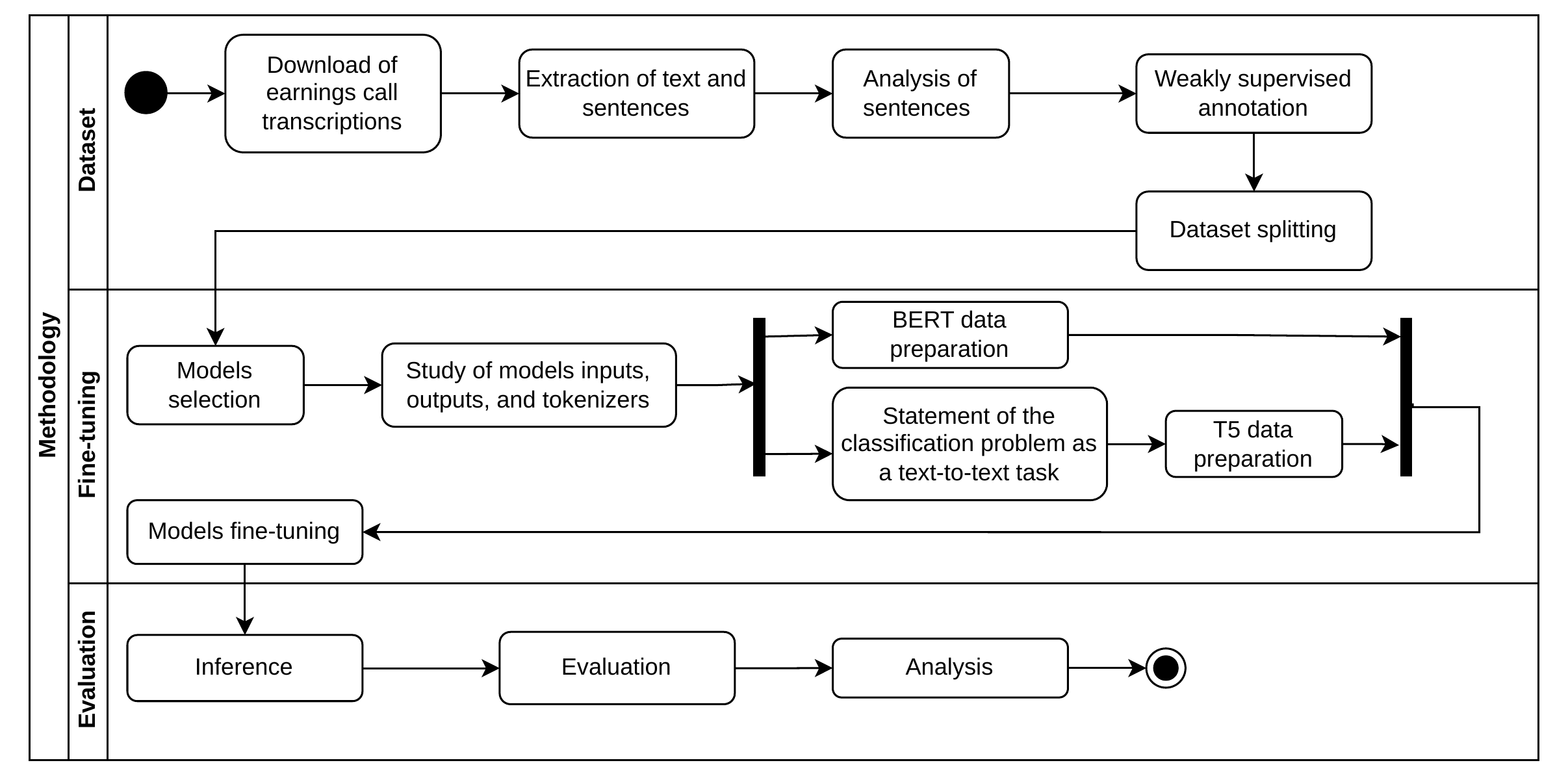}
    \caption{Overview of the main methodological steps}
    \label{fig:methodology}
\end{figure}

We conducted this work in three main steps, as shown in Fig. \ref{fig:methodology}. In the first step (Dataset), we constructed a dataset from earnings calls of Brazilian banks by extracting and annotating the sentences. Subsequently, in the Fine-tuning step, we fine-tuned four models. Finally, we analyzed their inference results in the Evaluation step. These steps are outlined in the following subsections. The scripts and datasets are available in a public repository\footnote{https://github.com/rsabilio/NerEval-BrazilianCorporateTranscripts}.

\subsection{Dataset development}

Earnings calls are conferences held by publicly traded companies to present their financial performance and results to stakeholders, including investors, analysts, and the media. These calls typically feature presentations by the company's executives, followed by a Question \& Answer session where analysts and investors can ask questions. The discussions often cover key financial statements such as the balance sheet, income statement, and cash flow statement. Some companies choose to transcribe these calls and make the transcripts publicly available. These transcripts are particularly interesting for study because they capture the nuances and informalities of spoken language, which can differ significantly from written text. This variability presents challenges for automatic information extraction. For example, the phrase `O lucro foi de 900 milhões de reais'' (The profit was BRL 900 million) might be transcribed as ``O lucro foi R\$900 milhões'', ``O lucro foi de R\$ 900 MM'', or simply ``O lucro foi de 900 milhões de reais''. Given these challenges, achieving high performance in Named Entity Recognition (NER) tasks by language models is crucial.

The following subsections present the development of the Brazilian Financial NER Dataset (BraFiNER), which comprises the acquisition of the transcript files and the annotation of the extracted sentences.

\subsubsection{Download of PDF files of Earnings Call}

We reviewed the Comissão de Valores Mobiliários (CVM)\footnote{The Comissão de Valores Mobiliários (CVM), akin to the U.S. Securities and Exchange Commission (SEC), is responsible for the development, regulation, and supervision of the Brazilian Securities Market. The CVM's mission is to facilitate capital raising for companies, safeguard investor interests, and ensure comprehensive disclosure of information regarding issuers and their securities.} open data \cite{CVM2023} and identified 29 active banks. We then visited their Investor Relations (IR) websites to check for available earnings call transcripts. For banks providing the PDF files, we compiled a list including the bank’s name, ticker (code of negotiation on the Brazilian Stock Exchange), start year for transcripts, and IR website URL. Using this list, we developed a crawler to download the PDF files published up to the second quarter of 2023.

Table \ref{tab:selected_banks} presents the ten selected banks along with their name, ticker, the year when the bank started providing the transcripts, and the majority ownership indicating whether the bank is controlled by the government or by private individuals or entities. We observe that two banks are controlled by the government: Banco do Brasil (Federal Government) and Banco do Estado do Rio Grande do Sul (State of Rio Grande do Sul), while eight are private banks. Notably, Banco do Brasil was the first bank to provide its transcripts in 2006. It is important to mention that, as of June 2024, the publication of earnings call transcripts is not mandatory. Hence, while other banks may provide video or audio of the earnings calls, some of the selected banks may not provide transcripts for specific quarters.

\begin{table}[ht]
    \centering
    \caption{Selected banks}
    \label{tab:selected_banks}
    \begin{tabular}{lccc}
        \hline
        Name                              & Ticker & Year & Majority       \\ 
                                          &        &      & ownership      \\
        \hline
        Banco ABC Brasil                  & ABCB   & 2008 & Private        \\
        Banco do Brasil                   & BBAS   & 2006 & Government     \\
        Banco Bradesco                    & BBDC   & 2015 & Private        \\
        Banco BMG                         & BMGB   & 2019 & Private        \\
        Banco PAN                         & BPAN   & 2012 & Private        \\
        Banco Estado do Rio Grande do Sul & BRSR   & 2007 & Government     \\
        Itaú Unibanco                     & ITUB   & 2011 & Private        \\
        Nu Holdings                       & NU     & 2021 & Private        \\
        Paraná Banco                      & PRBC   & 2009 & Private        \\
        Banco Santander Brasil            & SANB   & 2010 & Private        \\ 
        \hline
    \end{tabular}
\end{table}

\begin{figure}[!ht]
    \centering
    \includegraphics[scale=0.9]{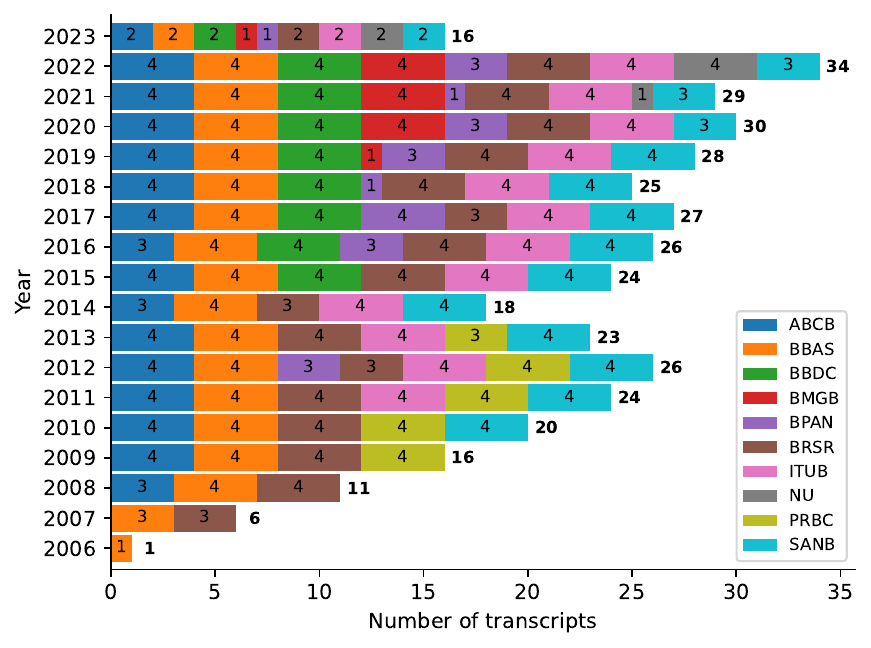}
    \caption{Number of transcripts per year and bank}
    \label{fig:number_of_transcripts_per_year_bank}
\end{figure}

With the crawler, we acquired 384 files published between 2006 and 2023, spanning 18 years. The distribution of transcripts per year and bank is visualized in Fig. \ref{fig:number_of_transcripts_per_year_bank}. For instance, in 2022, we obtained 34 transcripts from 9 banks, whereas only one was available from one bank in 2006. Such variations arise due to the diverse starting points for publication among the banks, occurring in different epochs, whether in a specific year or quarter.

After the transcripts download, we extracted the sentences from the PDF files using the PDFPlumber\footnote{https://github.com/jsvine/pdfplumber} v0.10.3 package. Subsequently, we employed NLTK\footnote{https://www.nltk.org/} v3.8 for sentence tokenization. Finally, all segmented sentences were compiled and stored in a CSV file for further processing and annotation. At the end of this process, we obtained 118,411 sentences. We analyzed those sentences observing their length regarding the number of words without punctuation and sentence duplicates.

We observed sentences ranging from 1 to 262 words in length. We filtered 9,103 sentences with up to 4 words, and upon analysis, determined that these short sentences, primarily greetings or phrases such as ``Por favor, aguardem.'' (``Please, wait.''), ``Obrigado.'' (``Thank you.''), and ``Muito obrigado a todos.'' (``Thank you all.''), would not significantly contribute to the domain. Additionally, we identified 10,833 duplicated sentences, including shorter and longer sentences such as ``Ótimo.'' (``Great.'') and ``A sessão de perguntas e respostas está encerrada.'' (``The question and answer session is closed.''). To enhance the quality of the dataset for fine-tuning models, we excluded both duplicated sentences and those with fewer than 4 words.

\begin{figure}[!ht]
    \centering
    \includegraphics[scale=0.65]{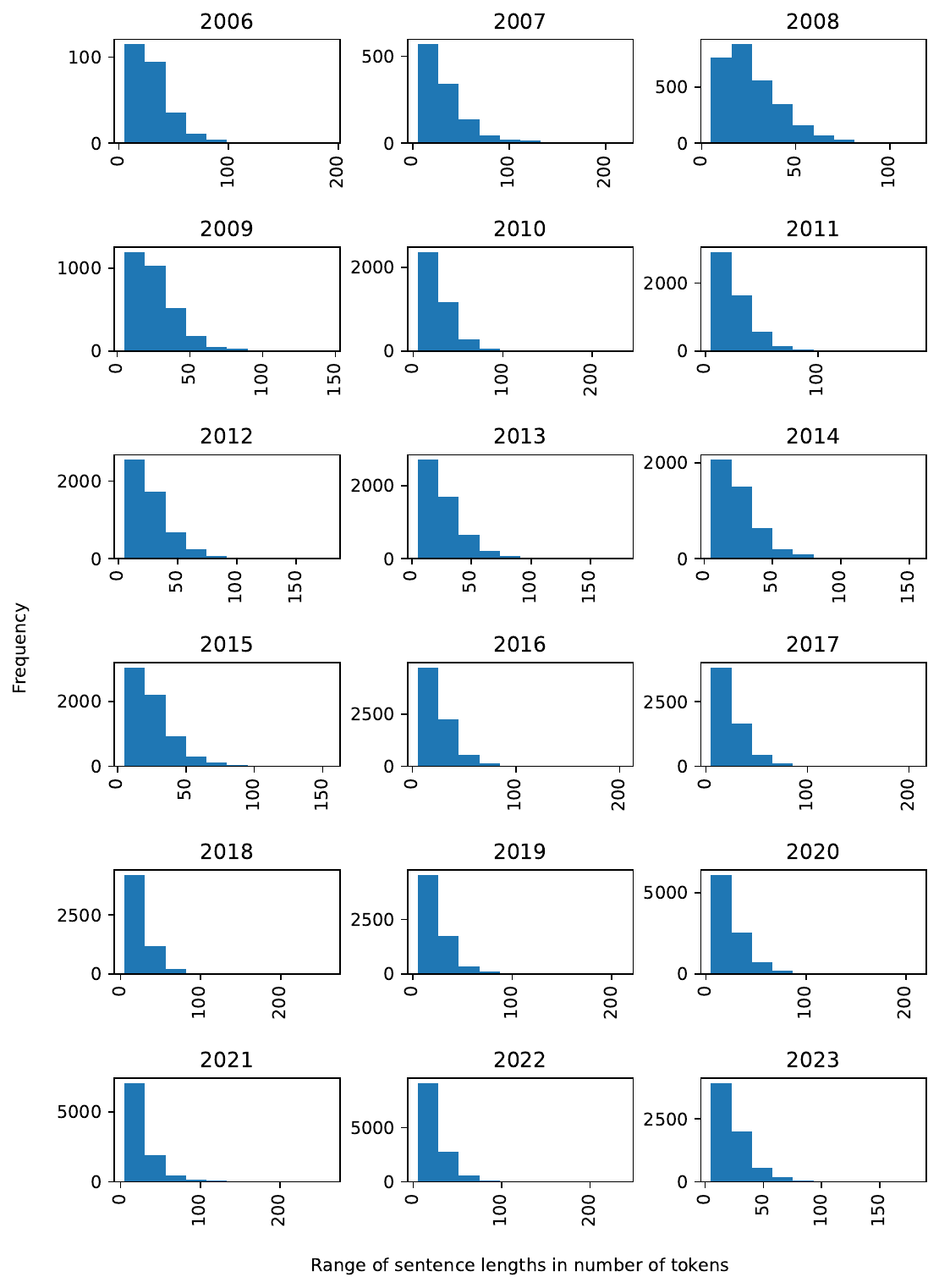}
    \caption{Frequency distribution of sentence lengths per year}
    \label{fig:freq_distribution_per_year}
\end{figure}

After this step, we amassed a corpus comprising 103,340 sentences, totaling 2,575,053 words. To elucidate the distribution of sentence lengths across different temporal segments, we plotted histograms (Fig. \ref{fig:freq_distribution_per_year}) with sentence length on the x-axis and the corresponding frequency of sentences on the y-axis, for each calendar year. For instance, our analysis of transcripts from 2022, totaling 34, revealed a spectrum of sentence lengths ranging from 5 to 236 words. Notably, within this range, a significant portion of 9,146 sentences exhibited lengths between 5 and 28 words.

Observing Fig. \ref{fig:freq_distribution_per_year}, it is evident that the shapes of the distributions are nearly identical, exhibiting positive skew and indicating that the majority of sentences have a lower number of words. According to our analysis, 75\% of sentences contain at least 32 words and the overall mean and standard deviation are 24.9 and 17.3, respectively.

\subsubsection{Data annotation}

We used a weakly supervised approach based on regular expressions, heuristics, and gazetteers \cite{Lison2021} to annotate the entities present in the extracted sentences. As an initial step, we defined a list of 23 labels, encompassing both generic labels, such as PERCENTUAL (Percentage), COMPANY, MONEY, YEAR, and ORG, as well as domain-specific labels including CARTEIRA (Portfolio), CLIENTE (Client), RISCO (Risk), PROVISAO (Reserve), DES\-PE\-SA (Expense), PRODUTO (Product), CON\-DI\-COES\_MA\-CRO\-E\-CO\-NO\-MI\-CAS (Macroeconomic Conditions), IN\-DI\-CA\-DOR\_LI\-QUI\-DEZ (Liquidity Indicator), IN\-DI\-CA\-DOR\_E\-FI\-CI\-EN\-CI\-A (Efficiency Indicator), QUARTER, LU\-CRO (Prof\-it), BA\-LAN\-CO\_PA\-TRI\-MO\-NI\-AL (Balance Sheet), SE\-MES\-TER, IN\-DI\-CA\-DOR\_REN\-TA\-BI\-LI\-DA\-DE (Profitability Indicator), RE\-CEI\-TA (Revenue), PRO\-VEN\-TO (Dividends), IN\-DI\-CA\-DOR\-\_VALUATION (Valuation Indicator), and RE\-SUL\-TA\-DOS (Results). The incorporation of these domain-specific labels distinguishes our approach from related work that typically relies on generic labels, even when dealing with datasets based on financial texts.

From that list, we developed a script using the Skweak package v0.3.3  \cite{Lison2021} to recognize those entities and annotate them. Skweak provides mechanisms based on regular expressions, functions (heu\-ris\-tics), and gazetteers. We implemented regular expressions to annotate year, quarter, and semester, functions to annotate percentage and money, and gazetteers to annotate the other entities. At the end of the annotation process, Skweak used a Hidden Markov Model to aggregate different annotations for the same entity resulting in a JSON structure containing a collection of sentences and their named entities (entity, label, start and end positions).

After running Skweak, a total of 119,517 annotations were generated from 57,933 sentences, resulting in an average of 2 annotations per sentence. From the total of sentences, the 45,407 sentences without annotations were filtered and stored in a separate CSV file. The distribution of entities per label is detailed in Table \ref{tab:number_of_entities_per_label_per_dataset}, arranged in descending order. Notably, the label ``PERCENTUAL'' (Percentage) exhibited the highest frequency with 26,934 occurrences, whereas ``RESULTADO'' (Result) was the least frequent, with 81 instances.

\begin{table}[ht]
    \centering
    \caption{Number of entities per label per dataset - Train, Val, and Test}
    \label{tab:number_of_entities_per_label_per_dataset}
    \begin{tabular}{lcccc}
    \hline
        Label                      & Train & Val  & Test & Total \\
    \hline
        PERCENTUAL                 & 19413 & 4892 & 2629 & 26934 \\
        CARTEIRA                   & 10435 & 2692 & 1429 & 14556 \\
        COMPANY                    & 9293  & 2338 & 1276 & 12907 \\
        CLIENTE                    & 9136  & 2203 & 1291 & 12630 \\
        MONEY                      & 5969  & 1520 & 809  & 8298  \\
        RISCO                      & 5082  & 1268 & 732  & 7082  \\
        PROVISAO                   & 4576  & 1127 & 655  & 6358  \\
        DESPESA                    & 3828  & 995  & 527  & 5350  \\
        PRODUTO                    & 3129  & 776  & 420  & 4325  \\
        QUARTER                    & 2507  & 569  & 336  & 3412  \\
        CONDICOES\_MACROECONOMICAS & 2228  & 579  & 308  & 3115  \\
        YEAR                       & 1921  & 440  & 303  & 2664  \\
        INDICADOR\_LIQUIDEZ        & 1629  & 394  & 239  & 2262  \\
        INDICADOR\_EFICIENCIA      & 1423  & 337  & 173  & 1933  \\
        RECEITA                    & 1321  & 299  & 167  & 1787  \\
        BALANCO\_PATRIMONIAL       & 837   & 220  & 118  & 1175  \\
        SEMESTER                   & 786   & 193  & 111  & 1090  \\
        LUCRO                      & 752   & 214  & 125  & 1091  \\
        ORG                        & 539   & 136  & 91   & 766   \\
        INDICADOR\_RENTABILIDADE   & 534   & 147  & 96   & 777   \\
        PROVENTO                   & 487   & 123  & 60   & 670   \\
        INDICADOR\_VALUATION       & 191   & 39   & 24   & 254   \\
        RESULTADO                  & 62    & 7    & 12   & 81    \\
    \hline
    \end{tabular}
\end{table}

We divided the 57,933 sentences into three datasets following the proportions: Train (70\% - 41,711 sentences), Validation (Val, 20\% - 10,428 sentences), and Test (10\% - 5,794 sentences). The Train and Validation datasets were utilized during the fine-tuning process, while the Test dataset was employed for evaluation purposes. The number of entities per label and dataset can also be observed in Table \ref{tab:number_of_entities_per_label_per_dataset}. For instance, the Train dataset has 19,413 PERCENTUAL, 752 LUCRO, and 62 RESULTADO entities. Despite the splitting process considering sentences rather than labels/entities, we can observe that all three datasets contain entities for every label.

\subsection{Models fine-tuning}
\label{sec:methodology:subsec:models-fine-tuning}

We searched on the Hugging Face platform for models provided by research groups or companies, specifically focusing on those pre-trained either exclusively in Portuguese or on multilingual datasets. We identified BERTimbau and PTT5 as models pre-trained in Portuguese, and mBERT and mT5 as multilingual models. While reviewing the available versions on Hugging Face, we opted for versions recommended by the original authors and with a comparable number of parameters to facilitate result comparison. Additionally, we included the smallest versions due to infrastructure constraints, despite the expectation of better results with larger models \cite{Devlin2019, Raffel2020}.

Table \ref{tab:meth:selected_models} shows the selected models and versions. For instance, BERT\-im\-bau is a BERT-based model pre-trained exclusively in Portuguese and we selected the Base version that has 110M of parameters. We can notice that BERTimbau and mBERT have the same number of parameters (110M), whereas PTT5 and mT5 differ. Initially, we experimented PTT5-small (60M parameters) to align with the BERT-based models, but it exhibited poor performance. Following the authors' recommendation, we selected the Base version with 220M parameters\footnote{https://github.com/unicamp-dl/PTT5}. 

\begin{table}
    \centering
    \caption{Selected models}
    \label{tab:meth:selected_models}
    \begin{tabular}{ccccc}
    \hline
         Architecture & Model     & Language     & Version & \#Parameters \\
    \hline
         BERT         & BERTimbau & Portuguese   & Base    & 110M         \\
         BERT         & mBERT     & Multilingual & Base    & 110M         \\
         T5           & PTT5      & Portuguese   & Base    & 220M         \\
         T5           & mT5       & Multilingual & Small   & 300M         \\
    \hline
    \end{tabular}
\end{table}

We studied the models' tokenizers and their inputs and outputs. After that, we developed scripts using Google Colaboratory (Colab) platform, Python v3.10, and Hugging Face libraries such as Transformers v4.35. We prepared the data developing different scripts for each architecture due to its characteristics. For example, while BERT-based models expect a sentence and a sequence of labels as inputs, T5-based models expect a sentence and an expected text (text that would be generated). For both architectures, we used the Beginning, Inside, and Outside (BIO) tagging scheme \cite{Ramshaw1995-BIO}.

In the BIO tagging scheme, each token in a sentence is assigned a tag. The tag ``O'' is assigned to tokens that are not part of any named entity. Conversely, the tags ``B'' and ``I'' are used for tokens that represent the beginning and continuation of a named entity, respectively. For instance, consider the label ``LUCRO'' (Profit), the entity ``lucro líquido'' (net income), and the sentence ``O lucro líquido aumentou'' (The net income increased). After tokenizing the sentence by whitespace, we obtain the tokens: ``O'', ``lucro'', ``líquido'', and ``aumentou''. Applying the BIO scheme results in the following sequence of tags: ``O'', ``B-LUCRO'', ``I-LUCRO'', ``O''. Here, ``O'' and ``aumentou'' are tagged as outside the entity with the tag ``O'', while ``lucro'' and ``líquido'' form the entity, with ``lucro'' marked as the beginning (B-LUCRO) and ``líquido'' as part of the entity (I-LUCRO). 

To fine-tune the BERT models, it was necessary to assign an integer to each label. Hence, we sorted the 23 labels in alphabetical order and created tags following the BIO scheme. For example, tags such as B-BA\-LAN\-CO\_PA\-TRI\-MO\-NI\-AL and I-BA\-LAN\-CO\_PA\-TRI\-MO\-NI\-AL were generated for the label BA\-LAN\-CO\_PA\-TRI\-MO\-NI\-AL. Sequential numbers were then assigned to each tag, where 0 corresponds to ``O'', 1 corresponds to B-BA\-LAN\-CO\_PA\-TRI\-MO\-NI\-AL, 2 corresponds to I-BA\-LAN\-CO\_PA\-TRI\-MO\-NI\-AL, and so forth. This list was saved in a JSON file and utilized in both the fine-tuning and evaluation steps.

The T5 model transforms text-based language problems into a text-to-text format \cite{Raffel2020}. Consequently, the token classification problem must be represented in a text-to-text format. In token classification, each token in a sentence is classified individually. For instance, a BERT model fine-tuned for NER processes a tokenized sentence and assigns a class to each token.

Various approaches to using T5 for NER are found in the literature, all representing the token classification task as a text generation task. Carmo et al. \cite{Carmo2020} input sentences into their T5-based model (PTT5) preceded by the instruction ``Recognize Entities'' and the model outputs the same sentence, but annotated with entity labels enclosed in brackets immediately following each entity. Wang et al. \cite{Wang2022instructionner} provide a T5 model with sentences preceded by an instruction and expect a list of sentences following the template ``entity is a/an type''.

In our approach, similar to Carmo et al. \cite{Carmo2020}, we input sentences into the model and expect the same sentence with entities and their labels. Unlike Carmo et al. \cite{Carmo2020} and Wang et al. \cite{Wang2022instructionner}, we do not use a preceding instruction because we fine-tuned our models for a single task. Wang et al. \cite{Wang2022instructionner} argue that their instruction helps the model identify labels and avoid omissions or additions. However, our testing revealed that this instruction led to longer inputs, increasing computational resource consumption during fine-tuning. Comparison of Precision, Recall, and macro F1-scores with and without the instruction favored omission of the instruction, resulting in inputs with fewer tokens, lower resource consumption, and higher metric values.

Therefore, in our approach, given a sentence S, the model outputs S', which is S with entities annotated following the template \texttt{[entity|label]} and replacing whitespaces with underscore for multi-word entities.

\begin{equation}
    S \rightarrow S'   
\end{equation}

For example, given $S$ as ``John lives in New York'', the model outputs $S'$ as ``[John|PERSON] lives in [New\_York|LOCATION]''. Hence, rather than classifying individual tokens or subwords, the model generates sentences with annotated entities.

The fine-tuning process was conducted using both Free and Pro (paid) versions of Google Colab. In the Free version, we used NVidia T4 GPU with 15 GB of memory, while in the Paid version, we used NVidia A100 GPU with 40 GB. It is noteworthy that the paid version was specifically required for fine-tuning the mT5 model.

\subsection{Evaluation of the models' results}
\label{subsection:evaluation_metrics}

We evaluated the models considering various aspects, including computational resource requirements for both fine-tuning and inference, performance metrics such as Precision, Recall, and macro F1-score, and error analysis based on the metrics from the Fifth Message Understanding Conference (MUC-5) \cite{Chinchor1993-muc}.

To calculate MUC-5 metrics, it's imperative to compare the labels assigned by the model with those assigned during the annotation step. The outcome is then classified into one of the following categories \cite{Chinchor1993-muc}: 

\begin{itemize}
    \item Correct (COR): The label assigned to a token by the model is equal to the label assigned at the annotation step;
    \item Incorrect (INC): The label assigned to a token by the model differs from the label assigned at the annotation step;
    \item Missing (MIS): There is a label assign at the annotation step, but the model did not assign any label;
    \item Spurious (SPU): The model assigned a label, but no label was assigned at the annotation step.
\end{itemize}

MUC-5 includes a category named ``Partial'', which is employed when the entity identified by the model partially matches the annotated entity in the dataset. However, our analysis focuses on the token level, and as a result, we opted not to use this metric. In consequence, we adapted the Primary and Secondary metrics, excluding the ``Partial'' metric from the equation.

We calculated Primary and Secondary metrics based on the categories, where lower values indicate better performance. The Primary metric ``Error per response fill'', adapted from MUC-5 \cite{Chinchor1993-muc}, is expressed by Eq. \ref{eq:error}. 

\begin{equation}
\label{eq:error}
    Error\ per \ response \ fill = \frac{(INC + MIS + SPU)}{(COR + INC + MIS + SPU)}
\end{equation}

The adapted Secondary metrics Undergeneration, Overgeneration, and Substitution are expressed by Eq. \ref{eq:undergeneration}, Eq. \ref{eq:overgeneration}, and Eq. \ref{eq:substitution}.

\begin{equation}
\label{eq:undergeneration}
    Undergeneration = \frac{MIS}{COR + INC + MIS}
\end{equation}

\begin{equation}
\label{eq:overgeneration}
    Overgeneration  = \frac{SPU}{COR + INC + SPU}  
\end{equation}

\begin{equation}
\label{eq:substitution}
    Substitution    = \frac{INC}{COR + INC}  
\end{equation}

Observing Eq. \ref{eq:error}, we can note that MUC-5 considers Spurious (SPU) as an error. However, we expect that the model classifies more tokens than the annotation process, given its capacity to learn. Consequently, we manually verified each entry classified as INC, MIS, and SPU to determine whether the model's assigned label can be considered correct or not.

We also verified the statistical differences among the models using the Wilcoxon test and Friedman's test, followed by the Nemenyi test \cite{Demsar2006-WilcoxonFriedmanNemenyi, Marozzi2014-Friedman-Ftest}. For Friedman's test, since we have a small number of subsets and classifiers, we calculated both  ${\chi}^2$ (Q) and F statistics  \cite{Demsar2006-WilcoxonFriedmanNemenyi, Marozzi2014-Friedman-Ftest}. Our null hypothesis ($H_0$) for the tests was that there is no statistical difference among the models at a significance level of 5\% ($\alpha = 0.05$), while our alternative hypothesis ($H_1$) was that differences do exist among the models. To perform the tests, we split the test dataset into five subsets, each of which was evaluated with every model. Based on the collected values, we calculated the statistical tests using the packages Pingouin v0.5.4  \cite{Vallat2018-pingouin} and SciPy v1.11.4 \cite{SciPy2020}.

\section{Experiments}
\label{sec:results}

With the annotated dataset, we performed the fine-tuning process for the BERT-based models BERTimbau and mBERT, as well as for the T5-based models PTT5 and mT5. This section outlines each fine-tuning process, detailing the data input formatting, characteristics of the datasets, definition of model hyperparameters, and key metrics of the fine-tuning process. These metrics include the GPU utilized, memory and time consumption during fine-tuning, and performance metrics.

\subsection{BERT-based models fine-tuning}
\label{subsection:bert_models_fine_tuning}

For BERT-based models, the datasets Train, Validation, and Test preparation involved the tokenization of the sentences into tokens, labeling in the BIO tagging scheme, tokenization of tokens into subwords, and alignment of labels to correspond to the subwords. Ultimately, we obtained lists of ``input\_ids'' and ``labels''. The processed datasets were saved in Hugging Face DatasetDict\footnote{https://huggingface.co/docs/datasets/index} structure for use in the fine-tuning.

To illustrate the preparation process, let's consider the sentence ``O lucro líquido do Santander aumentou'' (Santander's net profit increased) as the input, and in the annotation step, Skweak generated the following data as output: [[2, 15, ``lucro líquido'', ``LUCRO''], [19, 28, ``Santander'', ``COMPANY'']]. In this example, we have two entities: i) ``lucro líquido'', starting at position 2, ending at 15, and labeled as LUCRO; and ii) ``Santander'', starting at position 19, ending at 28, and labeled as COMPANY. With these data, the preparation process follows these two main steps:

\begin{itemize}
    \item Step 1: The sentence and its annotations were processed, resulting in a dictionary that maps tokens to their corresponding Named Entity Recognition (NER) tags, as illustrated below:
\end{itemize}

\begin{verbatim}
tokens: [[`O',`lucro',`líquido',`do',`Santander',`aumentou']],
ner_tags: [[0, 21, 22, 0, 7, 0]]
\end{verbatim}

In the list of ner\_tags, the numbers 21 and 22 correspond to the tags B-LUCRO and I-LUCRO, and the number 7 corresponds to the tag B-COMPANY.

\begin{itemize}
    \item Step 2: The tokens within the sentences were tokenized using the respective model's tokenizer, and the corresponding labels were adjusted accordingly.
\end{itemize}

For instance, at the end of this step, using BERTimbau's tokenizer, we have the following ``labels'' and ``input\_ids'' for the sentence ``O lucro líquido do Santander aumentou'':

\begin{verbatim}
labels: [[-100, 0, 21, 22, 0, 7, 7, 0, -100]]
input_ids: [[101,231,14699,10203,171,1838,1217,8231,102]]
\end{verbatim}

The ``input\_ids'' correspond to the subwords:

\begin{verbatim}
[`[CLS]',`O',`lucro',`líquido',`do',`Santa',`\#\#nder',
 `aumentou',`[SEP]']
\end{verbatim}

Notably, the word ``Santander'' was split into two subwords, ``Santa'' and ``\#\#nder,'' resulting in two instances of the number 7 (B-COMPANY) in the ``labels''. Additionally, ``-100'' in the ``labels'' corresponds to the special tokens ``[CLS]'' and ``[SEP]'' (numbers 101 and 102 present in ``input\_ids'').

The aforementioned steps were executed for every sentence in each dataset (Train, Val, and Test). Subsequently, we examined the vocabulary size of each model, calculated the number of tokens (words and punctuation), determined the number of subwords generated by the tokenizer, and assessed the ratio between subwords and tokens. The resulting data is presented in Table \ref{tab:proportion_subwords_tokens_bert}. For instance, BERTimbau has 29,794 subwords in its vocabulary and the calculated ratio between Subwords and Tokens was 1.19. This ratio means that it was generated 1.19 subword by the tokenizer for each token in a sentence.

\begin{table}[ht]
    \centering
    \caption{Measurements of vocabulary, subwords, and tokens - BERTimbau and mBERT}
    \label{tab:proportion_subwords_tokens_bert}
    \begin{tabular}{lcccc}
    \hline
        Model     & Vocab. size & \# Tokens & \# Subwords & Ratio   \\
    \hline
        BERTimbau & 29,794     & 1,996,940  & 2,381,471   & 1.19    \\
        mBERT     & 119,547    & 1,996,940  & 2,662,520   & 1.33    \\
    \hline
    \end{tabular}
\end{table}

After reviewing Table \ref{tab:proportion_subwords_tokens_bert}, it becomes apparent that the number of tokens remains consistent across models, as this calculation is model-independent. Particularly, mBERT produces a higher number of subwords compared to BERTimbau, with a ratio of 1.33 for mBERT and 1.19 for BERTimbau. This ratio plays a crucial role in determining model hyperparameters, as it helps estimate the number of tokens in the model's input. Additionally, it aids in decision-making processes regarding potential truncation of inputs or padding methods.

Regarding the vocabulary' size, as expected, the vocabulary of mBERT surpasses that of BERTimbau due to its extensive pre-training on a diverse corpus that includes 104 languages, including Portuguese. Following a detailed analysis, we identified a shared set of 14,322 subwords between BERTimbau and mBERT vocabularies. The higher number of subwords generated by the mBERT Tokenizer can be attributed to the relatively fewer subwords in Portuguese compared to BERTimbau.

Upon analyzing the sentence lengths in terms of the number of subwords, we noted that the longest sentence in mBERT comprised 379 subwords, while in BERTimbau, the maximum was 324 subwords. Based on this observation, we concluded that neither of the BERT models would require truncation of any sentences. Consequently, we proceeded with the fine-tuning of both BERTimbau and mBERT, utilizing batch sizes of 16, 2 epochs, and a learning rate of $5\cdot10^{-5}$ (the default value of Hugging Face Transformers TrainingArguments\footnote{https://huggingface.co/docs/transformers/main\_classes/trainer}) for each model. These hyperparameter values align with those outlined by Devlin et al. \cite{Devlin2019} and are deemed suitable for the available infrastructure, which includes an NVIDIA T4 GPU with 15 GB of memory.

Table \ref{tab:measurements_bertimbau_mbert_finetuning} summarizes the key measurements from the fine-tuning process of BERTimbau and mBERT, encompassing both the training and validation steps. These measurements include memory usage and time consumed (in minutes) throughout the entire process, as well as the Precision, Recall, and F1-score calculated from the validation dataset at the end of each training epoch. We can observe that mBERT requires a bit more memory and time for fine-tuning, even though they have the same number of parameters (110M). Despite this, both models exhibited slight differences in the Precision, Recall, and F1, with BERTimbau slightly outperforming mBERT in terms of F1-score. These data could aid in the decision on which model to choose, considering the requirements and the available infrastructure.

\begin{table}[ht]
    \centering
    \caption{Results of BERTimbau and mBERT fine-tuning}
    \label{tab:measurements_bertimbau_mbert_finetuning}
    \begin{tabular}{lcccccc}
    \hline
        Model     & GPU  & Memory   & Time   & Precision & Recall & F1     \\
                  &      & (GB)  & (min)  &           &        &        \\
    \hline
        BERTimbau & T4   & 11.2  & 14     & 0.9970    & 0.9985 & \textbf{0.9978}  \\
        mBERT     & T4   & 12.4  & 17     & 0.9962    & 0.9984 & 0.9973  \\
    \hline
    \end{tabular}
\end{table}

\subsection{T5-based models fine-tuning}

For T5-based models, each occurrence in every dataset underwent processing, involving the identification and annotation of each entity in the target sentence following our approach (cf. Section \ref{sec:methodology:subsec:models-fine-tuning}), and the tokenization of both sentences (input) and target (label). Ultimately, similar to BERT models (cf. Section \ref{subsection:bert_models_fine_tuning}), we obtained lists of ``input\_ids'' and ``labels'' and saved the processed datasets in the Hugging Face DatasetDict structure for use in fine-tuning.

To illustrate the process, consider the sentence and annotations presented in the BERT-models fine-tuning section (cf. Section \ref{subsection:bert_models_fine_tuning}). We performed these steps to prepare the data for T5-models fine-tuning:

\begin{itemize}
    \item Step 1: The sentence and its annotations were processed, resulting in a dictionary that maps the sentence input to its target sentence, as illustrated below:
\end{itemize}

\begin{verbatim}
input: [`O lucro líquido do Santander aumentou']
target: [`O [lucro_líquido|LUCRO] do [Santander|COMPANY] 
          aumentou']
\end{verbatim}

In this example, the entity ``lucro líquido'' consists of two words, which are joined by an underscore for processing. It is important to note that the BIO tagging scheme was not applied at this step. However, during the metric calculation phase, the generated sentence was split using whitespaces (with the underscore preventing the splitting of the entity). After generating the list of tokens, each token was iterated over, and the tag ``O'' was assigned to tokens that did not match the pattern \texttt{[entity|label]}. For tokens that matched the pattern, the BIO tags were applied. This same process was then applied to the target sentence. The resulting lists of BIO tags were then used to calculate the performance metrics.

\begin{itemize}
    \item Step 2: The input and target sentences were tokenized using the respective model's tokenizer
\end{itemize}

As a result of this step, using PTT5's tokenizer, we have the following ``labels'' and ``input\_ids'':

\begin{verbatim}

labels: [[28,1322,1034,2320,2005,2936,717,53,2847,503,10219,
          9508,1035,10,1322,8606,2573,2847,15974,388,6573,
          1995,1035,5999,1]],
input_ids: [[28, 12226, 5391, 10, 443, 2573, 5999, 1]]
\end{verbatim}

The ``input\_ids'' correspond to these subwords:

\begin{verbatim}
[`_O', `_lucro', `_líquido', `_do', `_Santa', `nder', 
`_aumentou', `</s>']
\end{verbatim}

Note that each number in ``labels'' and in ``input\_ids'' corresponds to a subword. Hence, after tokenization, we obtained more subwords in the target sentence (label) due to the addition of the annotation.

BERT and T5 use different tokenizers: WordPiece and SentencePiece, respectively. These tokenizers employ their own strategies for tokenization. Consequently, we can observe an underscore preceding each word (replacing spaces) in the T5 Tokenizer and ``\#\#'' indicating subwords in the BERT Tokenizer, as in ``Santa'' and ``\#\#nder''. Additionally, while the BERT Tokenizer uses the special tokens ``[CLS]'' and ``[SEP]'' to delimit the sentence, the T5 Tokenizer indicates the end of the sentence by $</s>$.

As in the BERT-models fine-tuning (cf. Section \ref{subsection:bert_models_fine_tuning}), we evaluated the sizes of the models' vocabularies (Vocab.), measured the number of tokens and subwords, and calculated the ratio between subwords and tokens for both input and target sentences. Table \ref{tab:proportion_subwords_tokens_t5} summarizes the measurements for each model. For example, the PTT5 vocabulary has 32,100 subwords, and the ratios between subwords and tokens for both input and target are 1.16 and 1.53, respectively. The number of tokens for both models was: a) Input: 2,004,673; and b) Target: 2,228,847, since the calculation is model-independent.

Analyzing Table \ref{tab:proportion_subwords_tokens_t5}, we observed that the vocabulary size of mT5 is nearly 7.8 times larger than that of PTT5. Upon calculating the intersection between the vocabularies, we found that there are 14,459 subwords in common, which is nearly the same number as the intersection between BERTimbau and mBERT. Further analysis of the ratios revealed an increase from input to target sentences for both models due to tokens added by the annotation. Additionally, we confirmed that the maximum number of subwords in input sentences was 324, while in target sentences, it reached 509 subwords for PTT5. For mT5, the corresponding maximum numbers were 480 and 511 for input and target sentences, respectively. Consequently, we set the hyperparameters of the models to 512 to avoid truncation and save computational resources.

\begin{table}[ht]
    \centering
    \caption{Measurements of vocabulary, subwords, and tokens - PTT5 and mT5}
    \label{tab:proportion_subwords_tokens_t5}
    \begin{tabular}{lccccc}
        \hline
        \multicolumn{1}{c}{Model} & Vocab.   & \multicolumn{2}{c}{Input} & \multicolumn{2}{c}{Target} \\
                                  & size    & \#Subwords     & Ratio    & \#Subwords   & Ratio     \\ \hline
        PTT5                      & 32,100  & 2,329,073      & 1.16     & 3,410,337    & 1.53      \\
        mT5                       & 250,100 & 3,165,279      & 1.58     & 3,785,740    & 1.70      \\ \hline
    \end{tabular}
\end{table}

With the dataset prepared, we proceeded with the fine-tuning process of PTT5 using both T4 (similar to the BERT models) and A100 GPUs. In contrast, the fine-tuning of mT5 utilized an A100 GPU, as this model required more memory than that provided by the T4. Throughout all fine-tuning experiments, we maintained a consistent approach, employing 2 epochs and batches of size 8. The key measurements from the fine-tuning process of PTT5 and mT5 are summarized in Table \ref{tab:measurements_ptt5_mt5_finetuning}. For instance, upon examination of Table \ref{tab:measurements_ptt5_mt5_finetuning}, it is apparent that mT5 consumed 33.5 GB of memory during 107 minutes (1.8 hours) of fine-tuning and achieved an F1-score of 0.9926.

Table \ref{tab:measurements_ptt5_mt5_finetuning} reveals noteworthy changes in the fine-tuning process of PTT5 when transitioning from the T4 GPU to the A100 GPU. The shift resulted in an increase in memory consumption by nearly 1.6 GB, accompanied by a 30-minute reduction in time and a slight decrease in F1 from 0.9919 to 0.9908. A comparison between PTT5 and mT5 fine-tuned using the A100 GPU indicates that mT5, despite consuming more memory, performed better with a 43-minute reduction in time and an overall slightly higher F1-score. 

\begin{table}[ht]
    \centering
    \caption{Results of PTT5 and mT5 fine-tuning}
    \label{tab:measurements_ptt5_mt5_finetuning}
    \begin{tabular}{lcccccc}
    \hline
        Model   & GPU  & Memory  & Time  & Precision & Recall & F1     \\
                &      & (GB) & (min) &           &        &        \\
    \hline
        PTT5    & T4   & 14.2 & 180   & 0.9917    & 0.9920 & 0.9919  \\
        PTT5    & A100 & 15.8 & 150   & 0.9915    & 0.9901 & 0.9908  \\
        mT5     & A100 & 33.5 & 107   & 0.9922    & 0.9930 & \textbf{0.9926}  \\
    \hline
    \end{tabular}
\end{table}

Additionally to the measurements of PTT5, we observed how quickly the model was processing and updating its parameters based on the training data since we used two different GPUs. When using T4 GPU, we observed a rate of 3.5 iterations (passes through batches of training data) per second and a rate of 6.2 iterations per second when using A100. This indicated, as expected, that the A100 GPU was faster than the T4, reducing the time to fine-tune PTT5 by 30 minutes.

\section{Evaluation}

\begin{figure}[!ht]
    \centering
    \includegraphics[scale=0.8]{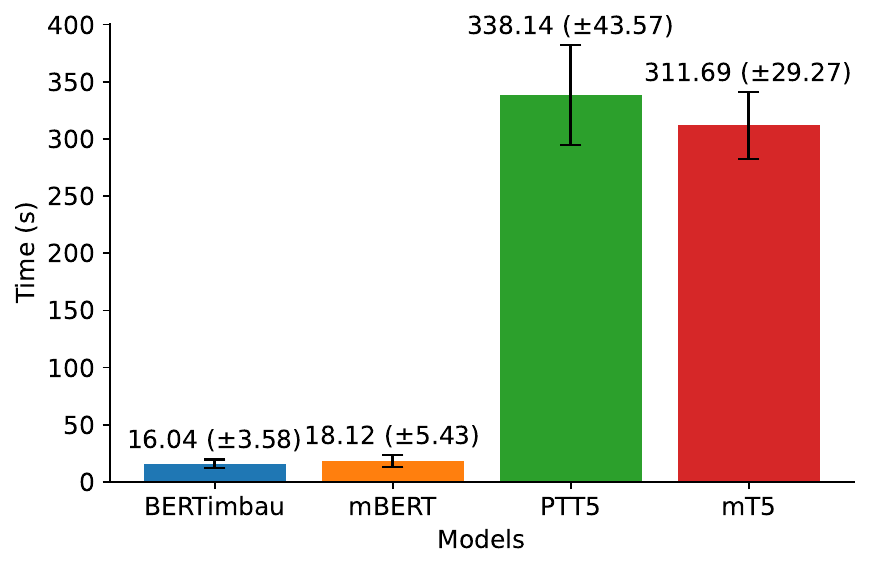}
    \caption{Average $\pm$ standard deviation of inference time per model (in seconds)}
    \label{fig:inference-results-time}
\end{figure}

The fine-tuned models were employed for inference on each subset of the Test dataset using batches of size 16. The mean and standard deviation (std) of the time utilized during the inference step are presented in Fig. \ref{fig:inference-results-time}. We can observe, in Fig. \ref{fig:inference-results-time}, a significant disparity between BERT and T5 models concerning the mean time for inference. While BERTimbau required a mean of 16.04 seconds ($\pm$ 3.58 s) to classify the subwords of the five subsets of the Test dataset, the PTT5 model required 28 minutes (an average of 338.14 $\pm$ 43.57 seconds) to generate the annotated sentences. Regarding memory usage, we observed that BERTimbau consumed 3.7 GB of memory, while mBERT and mT5 consumed 4.4 GB each. In contrast, PTT5 consumed 12.4 GB of memory, which is nearly three times the amount consumed by the other models.

\begin{figure}[!ht]
    \centering
    \includegraphics[scale=0.8]{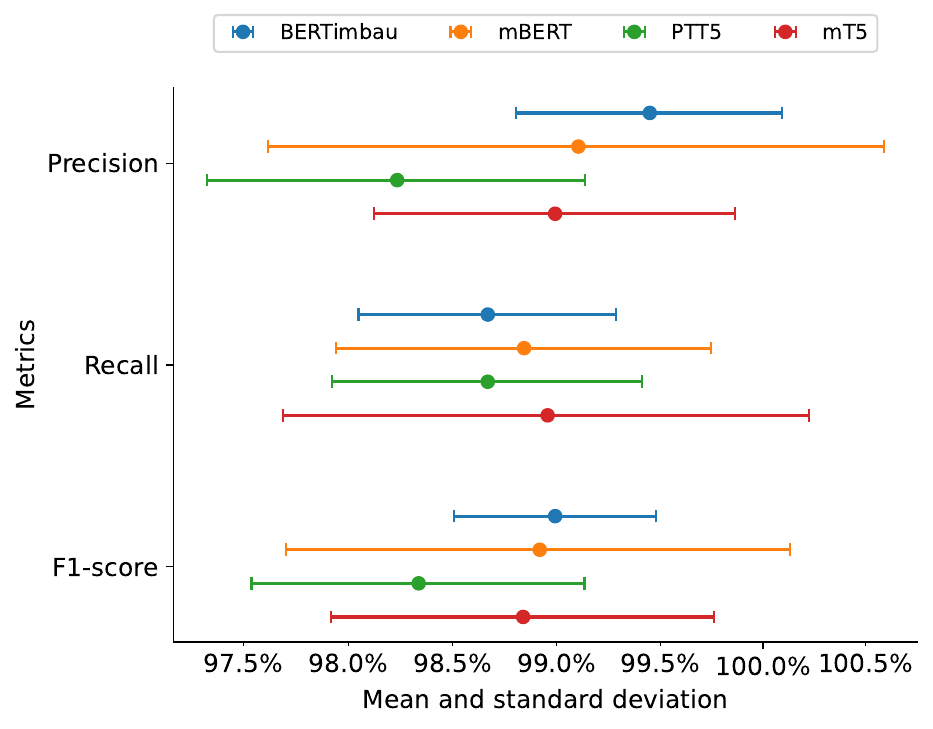}
    \caption{Performance of the models in inference step}
    \label{fig:inference-results-performance}
\end{figure}

Fig. \ref{fig:inference-results-performance} depicts the mean and standard deviation of Precision, Recall, and macro F1-score (F1) values in percentage (\%). Upon examining the performance of the models, we noted consistent values for these metrics across different GPUs for both PTT5 and mT5. Therefore, we only plotted the data for the T4 GPU for PTT5 and mT5.

Analyzing Fig. \ref{fig:inference-results-performance}, it is evident that BERTimbau achieved the highest average Precision (99.44\%) and macro F1-score (98.99\%) among all models, with the lowest standard deviations (0.64\% and 0.48\%, respectively), indicating more stable performance across evaluations. When comparing mono- and multilingual models, BERTimbau outperformed mBERT in terms of F1-score and the computational resources required for fine-tuning and inference. Conversely, among the T5 models, mT5 exhibited superior performance compared to PTT5 in terms of performance metrics. This difference may be attributed to the larger size of mT5, which has 80 million more parameters than PTT5. However, further investigation is needed to explore the specific characteristics of each architecture and pre-training process. Furthermore, the higher number of subwords in multilingual models seems to have no considerable impact on their performance when compared to monolingual models.

Despite the slight differences among the models, Friedman's test indicated a statistically significant difference (p-value $<$ 0.05) only for Precision (Q = 8.28, p-value = 0.04; F = 4.92, p-value = 0.02). Performing the Nemenyi test for Precision, we observed a statistically significant difference only between BERTimbau and PTT5 (p-value = 0.03). Analyzing Wilcoxon results, we observed that there is no statistically significant difference at the 5\% level between each pair of models for each metric.

Upon examining the measurements per class of all models (\ref{appendix:results_inference_per_class_model}), we observed consistently high values overall. However, we noted that BERTimbau and mBERT obtained the lowest F1-scores in the class RESULTADO, with 0.8878 and 0.8667, respectively. Analyzing the results for PTT5 and mT5, we also observed a low value of 0.9333 for both models in the class RESULTADO. Additionally, PTT5 and mT5 achieved their lowest F1-scores (0.9505 and 0.9353) in the class YEAR.

Verifying the entities labeled as RESULTADO in the gazetteers and annotated in the dataset, we found nested entities. For example, the entity ``Resultado das Operações de Seguros'' (Insurance Operations Result) is labeled as RESULTADO, and we also have ``Seguros'' as PRODUTO. Hence, Skweak annotated the entity as RESULTADO, but the models classified ``Resultado das Operações'' as RESULTADO and ``Seguros'' as PRODUTO. We also noted that RESULTADO is the class with the fewest number of entities in the dataset (Table \ref{tab:number_of_entities_per_label_per_dataset}). 

The nested entities and the number of entities are likely the cause of the low-performance scores for all models in the class RESULTADO. Nested entities are a particular challenge and were echoed by Shah et al. \cite{Shah2023}, who noted similar difficulties encountered by their model when handling nested entities.

After computing the performance metrics, we compared the labels assigned during the annotation step with those assigned by the model, following MUC-5 standards \cite{Chinchor1993-muc}. Additionally, we manually verified each classification, marking them with ``Yes'' when the model was correct and with ``No'' when the model was incorrect. The results of this evaluation are presented in Table \ref{tab:muc-5_class}. This table includes the class abbreviation, an indication of whether the model assigned the correct label, and the corresponding values for each model. For example, BERTimbau had 48 classifications marked as Missing (MIS), but upon manual verification, we found that the model was correct in 2 classifications, indicating that the annotations in the dataset were incorrect.

\begin{table}[ht]
\centering
\caption{Results of the manual verification}
\label{tab:muc-5_class}
\begin{tabular}{ccrrrr}
\hline
Metric                    & Correct?             & \multicolumn{1}{c}{BERTimbau} & \multicolumn{1}{c}{mBERT} & \multicolumn{1}{c}{PTT5} & \multicolumn{1}{c}{mT5} \\ \hline
COR                       &                      & 30058                         & 33114                     & 16362                    & 16418                   \\
MIS                       & No                   & 46                            & 44                        & 108                      & 61                      \\
                          & Yes                  & 2                             & 0                         & 8                        & 7                       \\
INC                       & No                   & 9                             & 14                        & 12                       & 8                       \\
                          & Yes                  & 0                             & 0                         & 0                        & 0                       \\
SPU                       & No                   & 12                            & 25                        & 9                        & 97                      \\
                          & Yes                  & 83                            & 95                        & 115                      & 70                      \\ 
\hline
   
\end{tabular}
\end{table}

We observe in Table \ref{tab:muc-5_class} that PTT5 generated more correct additional classifications (115) than the other models, followed by mBERT (95). According to MUC-5, the term ``Spurious'' is interpreted as an error because the model classified more tokens than those annotated in the dataset. However, it is important to note that we anticipated that the models, after fine-tuning, would be capable of identifying and classifying entities not annotated in the original dataset, resulting in additional correct predictions. Therefore, from our perspective, it is a positive scenario when a model generates correct additional classifications.

While manually analyzing the classifications generated by PTT5 and mT5, we observed instances of text repetition and alterations in the generated text. We further categorized the incorrectly generated text as either ``critical'' or ``non-critical''. Non-critical errors are related to minor changes in the text that do not alter the sentence's meaning, while critical errors involve changes in numerical values, words, and text repetition. An example of a critical error is the alteration of a numerical representation of money: from the original target \texttt{[R\$\_824,00\textbar MONEY]} to the generated \texttt{[R\$\_8924,00\textbar MONEY]}. In this case, the original entity was ``824,00'', and the model erroneously generated ``8924,00'', representing a nearly tenfold increase. Other examples include:

\begin{itemize}
    \item In the entity ``Receita Federal'' (the Brazilian Federal Revenue Service) the model changed ``Federal'' to ``fixa'' (fixed) generating ``receita fixa'' (fixed revenue);  
    \item In a sentence ending with the segment ``\texttt{[R\$\_10,8\_bilhões\textbar MONEY]} para acho que está em torno de \texttt{[R\$\_9,8\textbar MONEY]}'' (\texttt{[R\$\_10.8\_bil\-lion\textbar MONEY]} to I think it's around \texttt{[R\$9.8\textbar MONEY]}), the model altered ``10,8'' to ``8,8'' and repeated the entire segment eight times; and
    \item In the segment ``prejuízo foram \texttt{[0,08\%\textbar PERCENTUAL]}'' (losses were\newline\texttt{[0.08\%\textbar PERCENTAGE]}), ``0.08\%'' was changed to ``0.88\%'', an increase of eleven times.
\end{itemize}

\begin{figure}[ht]
    \centering
    \includegraphics[scale=0.7]{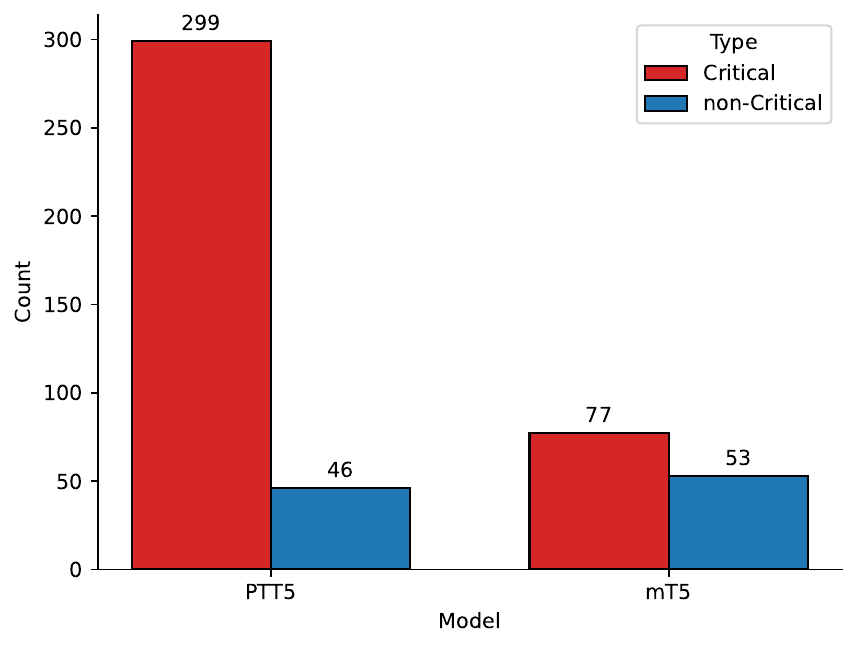}
    \caption{Number of critical and non-critical errors - PTT5 and mT5}
    \label{fig:number_of_critial_and_non-critical_errors_ptt5_mt5}
\end{figure}

Observing Figure \ref{fig:number_of_critial_and_non-critical_errors_ptt5_mt5}, it is evident that PTT5 generated 345 sentences with errors — 299 with critical errors and 46 with non-critical errors. This accounts for nearly 6\% of the Test dataset. In contrast, mT5 generated 130 sentences with errors, representing almost 2.3\% of the dataset — 77 with critical errors and 53 with non-critical errors.

To verify how similar the generated text was to the original target, we assessed the similarity between the original target sentences and their corresponding counterparts with errors. To do so, we employed the SequenceMatcher class from Python's ``difflib'' package\footnote{https://docs.python.org/3/library/difflib.html}. We chose this strategy because this algorithm compares pairs of sequences, aiming to identify the longest contiguous matching subsequences that do not contain elements such as blank lines or whitespaces. Utilizing the total number of elements in each sequence (T) and the number of matches (M), the ``ratio()'' method calculates the similarity as $2*M/T$, yielding a float in the range [0, 1].

Using the similarity ratios, we examined the frequency distribution of sentences categorized as critical or non-critical, along with their respective similarity ratios. For PTT5, among the 299 sentences with critical errors, an overwhelming majority (98\%) exhibited a similarity ratio in the range of 0.93 to 1.0, while 96\% of the sentences with non-critical errors had a similarity ratio between 0.97 and 1.0. In the case of mT5, out of the 77 sentences with critical errors, 70 (91\%) had a similarity ratio from 0.91 to 1.0, while among the 53 sentences with non-critical errors, 43 (87\%) had a similarity ratio ranging from 0.99 to 1.0.

These data indicate that, by employing our strategy to convert the token classification problem into a text generation problem, both models (PTT5 and mT5) tended to produce text with high similarity ratios to the original sentences, even in the presence of critical errors. This suggests that, despite these errors, the generated outputs retained a significant resemblance to the intended meaning of the original sentences. However, if precision is crucial, especially when dealing with exact numbers such as those related to money or percentages, relying solely on these models might be risky due to the potential for critical errors. We did not observe these issues with BERT-based models, as they output only the class for a given subword, whereas the T5 models generate the entire sentence.

Based on the data presented in Table \ref{tab:muc-5_class}, we calculated MUC-5 metrics for the five subsets of the Test dataset. Table \ref{tab:muc-5_metrics_values} presents the mean values of the metrics with their standard deviation. Analyzing these metrics, we observe that BERTimbau exhibits lower values for key metrics. Specifically, in terms of ``Error per response fill'', BERTimbau demonstrates correctness in more situations compared to the other models. Additionally, it excels in ``Overgeneration'', generating proportionally fewer incorrect spurious classifications, and in ``Substitution'', making fewer changes from correct to incorrect classifications. On the other hand, mBERT performs better in ``Undergeneration'', missing fewer classifications than the other models.

\begin{table}[ht]
\centering
\caption{Mean $\pm$ standard deviation values of adapted MUC-5 metrics}
\label{tab:muc-5_metrics_values}
\resizebox{\columnwidth}{!}{
\begin{tabular}{lrrrr}
\hline
Metric                  & \multicolumn{1}{c}{BERTimbau}  & \multicolumn{1}{c}{mBERT}   & \multicolumn{1}{c}{PTT5} & \multicolumn{1}{c}{mT5} \\

\hline

Error per response fill & \textbf{0.2212} ($\pm$0.08)    & 0.2503 ($\pm$0.10)          & 0.7775 ($\pm$0.14)       & 0.9893 ($\pm$0.88) \\
Overgeneration          & \textbf{0.0397} ($\pm$0.02)    & 0.0754 ($\pm$0.03)          & 0.0544 ($\pm$0.05)       & 0.5790 ($\pm$0.79) \\
Substitution            & \textbf{0.0299} ($\pm$0.02)    & 0.0422 ($\pm$0.04)          & 0.0730 ($\pm$0.05)       & 0.0484 ($\pm$0.05) \\
Undergeneration         & 0.1522 ($\pm$0.07)             & \textbf{0.1334} ($\pm$0.07) & 0.6563 ($\pm$0.14)       & 0.3696 ($\pm$0.16) \\

\hline

\end{tabular}
}
\end{table}

Regarding the Friedman's test for the metrics in Table \ref{tab:muc-5_metrics_values}, we found a statistically significant difference at the 5\% level for ``Error per response fill'' (Q = 12.12, p-value = 0.007; F = 16.83, p-value = 0.000337), ``Undergeneration'' (Q = 12.84, p-value = 0.005; F = 23.78, p-value = 0.000077), and ``Substitution'' (Q = 7.70, p-value = 0.0527; F = 4.21, p-value = 0.038). We calculated the Nemenyi test for these metrics and observed a statistically significant difference in ``Error per response fill'' between BERTimbau and PTT5 (p-value = 0.036). For ``Undergeneration'', the statistically significant difference appears between mBERT and PTT5 (p-value = 0.003). Despite Friedman's test indicating a statistically significant difference in F-statistic for ``Substitution'', the Nemenyi test only showed a p-value of 0.068 between BERTimbau and PTT5.

After calculating the Wilcoxon test for each metric, we observed p-values of 0.062 or higher. Therefore, we accept the null hypothesis: there is no statistically significant difference between the pairs of models.

Based on the statistical tests, we can conclude that BERTimbau exhibited superior performance among the models. We also observed that there are no statistically significant differences between mono and multi-lingual models (BERTimbau versus mBERT and PTT5 versus mT5).

\section{Conclusions}
\label{sec:conclusion}

In this study, we evaluated four Transformer-based models: BERTimbau and PTT5, which were exclusively pre-trained in Portuguese, along with the multilingual models mBERT and mT5, for the financial Named Entity Recognition (NER) task. We introduced BraFiNER, a novel dataset specifically designed for NER tasks in the Portuguese financial domain. This dataset includes 23 labels covering both generic and domain-specific categories, making it the first of its kind. 

Based on our evaluation, BERTimbau emerges as the preferred model due to its superior performance in both metrics and computational efficiency during inference. Statistical tests comparing mono- and multi-lingual models revealed no significant differences between them.

Through manual error analysis, we identified issues from the weakly supervised annotation process. While regular expressions and heuristics are helpful, they don't cover all cases, particularly with money units like millions and billions, where annotations were sometimes limited to currency symbols and numbers. We also observed difficulties with nested or multi-word entities and variations in entity spellings or formats, such as singular and plural forms. These limitations are inherent to weak supervision. Despite these challenges, the dataset remains valuable for further research.

Our approach to framing the token classification problem as a text generation task yielded promising results, notably with T5 models generating outputs closely resembling the original sentences. However, we encountered critical issues where the model altered monetary and percentage values, which are crucial in the financial domain. Despite these challenges, our strategy proved effective, as we fine-tuned models with fewer parameters and achieved higher F1-scores compared to related work \cite{Wang2022instructionner}.

For future work, we plan to expand the dataset with additional transcripts, as companies release them quarterly. Additionally, we aim to improve the dataset quality and revise the labels list for comprehensiveness and accuracy. Furthermore, we intend to explore prompt engineering techniques in generative models.

\bibliographystyle{elsarticle-num} 
\bibliography{refs-v2.1}





\appendix

\section{Results of inference using Test dataset per class and model}
\label{appendix:results_inference_per_class_model}

\begin{table}[ht]
\centering
\caption{Results per class - BERTimbau}
\label{tab:results_per_class_BERTimbau}
\resizebox{\columnwidth}{!}{
\begin{tabular}{lrrr}
\hline
                            & Precision       & Recall          & F1-score        \\      
\hline
BALANCO\_PATRIMONIAL        & 0.9938          & 0.9802          & 0.9867          \\          
CARTEIRA                    & 0.9996          & 0.9968          & 0.9982          \\          
CLIENTE                     & 1.0000          & 0.9993          & 0.9996          \\         
COMPANY                     & 0.9929          & 0.9976          & 0.9952          \\         
CONDICOES\_MACROECONOMICAS  & 0.9978          & 0.9946          & 0.9962          \\          
DESPESA                     & 1.0000          & 1.0000          & 1.0000          \\          
INDICADOR\_EFICIENCIA       & 1.0000          & 1.0000          & 1.0000          \\          
INDICADOR\_LIQUIDEZ         & 0.9978          & 1.0000          & 0.9989          \\          
INDICADOR\_RENTABILIDADE    & 0.9727          & 0.9943          & 0.9832          \\          
INDICADOR\_VALUATION        & 1.0000          & 0.9385          & 0.9636          \\           
LUCRO                       & 1.0000          & 1.0000          & 1.0000          \\          
MONEY                       & 0.9974          & 0.9892          & 0.9933          \\         
ORG                         & 1.0000          & 1.0000          & 1.0000          \\           
PERCENTUAL                  & 0.9979          & 0.9943          & 0.9961          \\         
PRODUTO                     & 1.0000          & 0.9913          & 0.9956          \\          
PROVENTO                    & 1.0000          & 1.0000          & 1.0000          \\          
PROVISAO                    & 0.9982          & 1.0000          & 0.9991          \\         
QUARTER                     & 1.0000          & 1.0000          & 1.0000          \\          
RECEITA                     & 0.9818          & 0.9818          & 0.9818          \\          
RESULTADO                   & 0.9500          & 0.8433          & 0.8878          \\           
RISCO                       & 0.9992          & 0.9973          & 0.9982          \\         
SEMESTER                    & 1.0000          & 0.9950          & 0.9975          \\          
YEAR                        & 0.9943          & 1.0000          & 0.9971          \\ 
\hline
macro avg                   & 0.9945          & 0.9867          & 0.9899          \\        
micro avg                   & 0.9977          & 0.9958          & 0.9968          \\        
weighted avg                & 0.9978          & 0.9958          & 0.9967          \\        
\hline

\end{tabular}
}
\end{table}

\begin{table}[ht]
\centering
\caption{Results per class - mBERT}
\label{tab:results_per_class_mBERT}
\resizebox{\columnwidth}{!}{
\begin{tabular}{lrrrr}
\hline
                             & Precision       & Recall          & F1-score        \\      
\hline
BALANCO\_PATRIMONIAL         & 1.0000          & 0.9771          & 0.9882          \\          
CARTEIRA                     & 0.9996          & 0.9976          & 0.9986          \\         
CLIENTE                      & 0.9985          & 0.9986          & 0.9986          \\         
COMPANY                      & 0.9920          & 0.9976          & 0.9948          \\         
CONDICOES\_MACROECONOMICAS   & 0.9987          & 0.9968          & 0.9978          \\          
DESPESA                      & 0.9994          & 1.0000          & 0.9997          \\         
INDICADOR\_EFICIENCIA        & 1.0000          & 0.9930          & 0.9964          \\          
INDICADOR\_LIQUIDEZ          & 0.9901          & 0.9952          & 0.9926          \\          
INDICADOR\_RENTABILIDADE     & 0.9910          & 0.9873          & 0.9891          \\          
INDICADOR\_VALUATION         & 1.0000          & 0.9500          & 0.9714          \\           
LUCRO                        & 1.0000          & 0.9913          & 0.9956          \\          
MONEY                        & 0.9961          & 0.9897          & 0.9929          \\         
ORG                          & 0.9923          & 0.9923          & 0.9923          \\           
PERCENTUAL                   & 0.9986          & 0.9949          & 0.9967          \\         
PRODUTO                      & 1.0000          & 0.9861          & 0.9929          \\          
PROVENTO                     & 1.0000          & 1.0000          & 1.0000          \\          
PROVISAO                     & 1.0000          & 0.9984          & 0.9992          \\         
QUARTER                      & 0.9931          & 0.9979          & 0.9955          \\          
RECEITA                      & 1.0000          & 1.0000          & 1.0000          \\          
RESULTADO                    & 0.8500          & 0.9000          & 0.8667          \\           
RISCO                        & 0.9976          & 0.9974          & 0.9975          \\         
SEMESTER                     & 1.0000          & 1.0000          & 1.0000          \\          
YEAR                         & 0.9968          & 0.9925          & 0.9946          \\          
\hline
macro avg                    & 0.9910          & 0.9884          & 0.9892          \\        
micro avg                    & 0.9978          & 0.9955          & 0.9966          \\        
weighted avg                 & 0.9979          & 0.9955          & 0.9966          \\        
\hline
\end{tabular}
}
\end{table}

\begin{table}[ht]
\centering
\caption{Results per class - PTT5 on T4 and A100 GPUs}
\label{tab:results_per_class_PTT5}
\resizebox{\columnwidth}{!}{
\begin{tabular}{lrrrr}
\hline
                             & Precision       & Recall          & F1-score        \\      
\hline
BALANCO\_PATRIMONIAL         & 1.0000          & 0.9802          & 0.9898          \\          
CARTEIRA                     & 0.9929          & 0.9916          & 0.9922          \\         
CLIENTE                      & 0.9985          & 0.9977          & 0.9981          \\         
COMPANY                      & 0.9855          & 0.9962          & 0.9908          \\         
CONDICOES\_MACROECONOMICAS   & 0.9903          & 0.9910          & 0.9905          \\          
DESPESA                      & 0.9981          & 0.9981          & 0.9981          \\          
INDICADOR\_EFICIENCIA        & 1.0000          & 1.0000          & 1.0000          \\          
INDICADOR\_LIQUIDEZ          & 0.9749          & 0.9831          & 0.9790          \\          
INDICADOR\_RENTABILIDADE     & 0.9549          & 0.9933          & 0.9724          \\           
INDICADOR\_VALUATION         & 1.0000          & 0.9278          & 0.9597          \\           
LUCRO                        & 1.0000          & 0.9913          & 0.9956          \\          
MONEY                        & 0.9887          & 0.9823          & 0.9855          \\          
ORG                          & 0.9746          & 1.0000          & 0.9869          \\           
PERCENTUAL                   & 0.9944          & 0.9943          & 0.9944          \\         
PRODUTO                      & 0.9884          & 0.9816          & 0.9850          \\          
PROVENTO                     & 1.0000          & 0.9846          & 0.9920          \\           
PROVISAO                     & 0.9982          & 0.9968          & 0.9975          \\          
QUARTER                      & 0.9910          & 0.9910          & 0.9910          \\          
RECEITA                      & 0.9548          & 0.9548          & 0.9548          \\          
RESULTADO                    & 0.9000          & 1.0000          & 0.9333          \\           
RISCO                        & 0.9987          & 0.9948          & 0.9967          \\          
SEMESTER                     & 0.9913          & 0.9778          & 0.9838          \\          
YEAR                         & 0.9185          & 0.9853          & 0.9505          \\          
\hline
macro avg                    & 0.9823          & 0.9867          & 0.9834          \\        
micro avg                    & 0.9905          & 0.9921          & 0.9913          \\        
weighted avg                 & 0.9908          & 0.9921          & 0.9914          \\        
\hline
\end{tabular}
}
\end{table}

\begin{table}[ht]
\centering
\caption{Results per class - mT5 on T4 and A100 GPUs}
\label{tab:results_per_class_mT5}
\resizebox{\columnwidth}{!}{
\begin{tabular}{lrrrr}
\hline
                             & Precision       & Recall          & F1-score        \\      
\hline
BALANCO\_PATRIMONIAL         & 1.0000          & 0.9889          & 0.9943          \\          
CARTEIRA                     & 0.9979          & 0.9972          & 0.9975          \\         
CLIENTE                      & 0.9985          & 0.9977          & 0.9981          \\         
COMPANY                      & 0.9892          & 0.9992          & 0.9942          \\         
CONDICOES\_MACROECONOMICAS   & 0.9967          & 0.9971          & 0.9969          \\          
DESPESA                      & 0.9945          & 0.9947          & 0.9946          \\          
INDICADOR\_EFICIENCIA        & 1.0000          & 1.0000          & 1.0000          \\          
INDICADOR\_LIQUIDEZ          & 0.9963          & 0.9963          & 0.9963          \\          
INDICADOR\_RENTABILIDADE     & 0.9716          & 0.9867          & 0.9783          \\           
INDICADOR\_VALUATION         & 1.0000          & 1.0000          & 1.0000          \\           
LUCRO                        & 1.0000          & 0.9913          & 0.9956          \\          
MONEY                        & 0.9820          & 0.9706          & 0.9762          \\          
ORG                          & 0.9889          & 1.0000          & 0.9943          \\           
PERCENTUAL                   & 0.9951          & 0.9883          & 0.9917          \\         
PRODUTO                      & 1.0000          & 0.9867          & 0.9933          \\          
PROVENTO                     & 0.9882          & 0.9882          & 0.9882          \\           
PROVISAO                     & 0.9946          & 0.9946          & 0.9946          \\          
QUARTER                      & 0.9934          & 0.9905          & 0.9919          \\          
RECEITA                      & 1.0000          & 1.0000          & 1.0000          \\          
RESULTADO                    & 0.9000          & 1.0000          & 0.9333          \\           
RISCO                        & 0.9961          & 0.9962          & 0.9961          \\          
SEMESTER                     & 1.0000          & 0.9846          & 0.9920          \\          
YEAR                         & 0.9851          & 0.9111          & 0.9353          \\          
\hline
macro avg                    & 0.9899          & 0.9896          & 0.9884          \\        
micro avg                    & 0.9942          & 0.9885          & 0.9913          \\        
weighted avg                 & 0.9943          & 0.9885          & 0.9907          \\        
\hline
\end{tabular}
}
\end{table}

\end{document}